\documentclass[11pt]{article}
\makeatletter

\newcounter{treecount}
\newcounter{branchcount}
\newsavebox{\parentbox}
\newsavebox{\treebox}
\newsavebox{\treeboxone}
\newsavebox{\treeboxtwo}
\newsavebox{\treeboxthree}
\newsavebox{\treeboxfour}
\newsavebox{\treeboxfive}
\newsavebox{\treeboxsix}
\newsavebox{\treeboxseven}
\newsavebox{\treeboxeight}
\newsavebox{\treeboxnine}
\newsavebox{\treeboxten}
\newsavebox{\treeboxeleven}
\newsavebox{\treeboxtwelve}
\newsavebox{\treeboxthirteen}
\newsavebox{\treeboxfourteen}
\newsavebox{\treeboxfifteen}
\newsavebox{\treeboxsixteen}
\newsavebox{\treeboxseventeen}
\newsavebox{\treeboxeighteen}
\newsavebox{\treeboxnineteen}
\newsavebox{\treeboxtwenty}
\newlength{\treeoffsetone}
\newlength{\treeoffsettwo}
\newlength{\treeoffsetthree}
\newlength{\treeoffsetfour}
\newlength{\treeoffsetfive}
\newlength{\treeoffsetsix}
\newlength{\treeoffsetseven}
\newlength{\treeoffseteight}
\newlength{\treeoffsetnine}
\newlength{\treeoffsetten}
\newlength{\treeoffseteleven}
\newlength{\treeoffsettwelve}
\newlength{\treeoffsetthirteen}
\newlength{\treeoffsetfourteen}
\newlength{\treeoffsetfifteen}
\newlength{\treeoffsetsixteen}
\newlength{\treeoffsetseventeen}
\newlength{\treeoffseteighteen}
\newlength{\treeoffsetnineteen}
\newlength{\treeoffsettwenty}

\newlength{\treeshiftone}
\newlength{\treeshifttwo}
\newlength{\treeshiftthree}
\newlength{\treeshiftfour}
\newlength{\treeshiftfive}
\newlength{\treeshiftsix}
\newlength{\treeshiftseven}
\newlength{\treeshifteight}
\newlength{\treeshiftnine}
\newlength{\treeshiftten}
\newlength{\treeshifteleven}
\newlength{\treeshifttwelve}
\newlength{\treeshiftthirteen}
\newlength{\treeshiftfourteen}
\newlength{\treeshiftfifteen}
\newlength{\treeshiftsixteen}
\newlength{\treeshiftseventeen}
\newlength{\treeshifteighteen}
\newlength{\treeshiftnineteen}
\newlength{\treeshifttwenty}
\newlength{\treewidthone}
\newlength{\treewidthtwo}
\newlength{\treewidththree}
\newlength{\treewidthfour}
\newlength{\treewidthfive}
\newlength{\treewidthsix}
\newlength{\treewidthseven}
\newlength{\treewidtheight}
\newlength{\treewidthnine}
\newlength{\treewidthten}
\newlength{\treewidtheleven}
\newlength{\treewidthtwelve}
\newlength{\treewidththirteen}
\newlength{\treewidthfourteen}
\newlength{\treewidthfifteen}
\newlength{\treewidthsixteen}
\newlength{\treewidthseventeen}
\newlength{\treewidtheighteen}
\newlength{\treewidthnineteen}
\newlength{\treewidthtwenty}
\newlength{\daughteroffsetone}
\newlength{\daughteroffsettwo}
\newlength{\daughteroffsetthree}
\newlength{\daughteroffsetfour}
\newlength{\branchwidthone}
\newlength{\branchwidthtwo}
\newlength{\branchwidththree}
\newlength{\branchwidthfour}
\newlength{\parentoffset}
\newlength{\treeoffset}
\newlength{\daughteroffset}
\newlength{\branchwidth}
\newlength{\parentwidth}
\newlength{\treewidth}
\newcommand{\ontop}[1]{\begin{tabular}{c}#1\end{tabular}}
\newcommand{\poptree}{%
\ifnum\value{treecount}=0\typeout{QobiTeX warning---Tree stack underflow}\fi%
\addtocounter{treecount}{-1}%
\setlength{\treeoffsettwo}{\treeoffsetthree}%
\setlength{\treeoffsetthree}{\treeoffsetfour}%
\setlength{\treeoffsetfour}{\treeoffsetfive}%
\setlength{\treeoffsetfive}{\treeoffsetsix}%
\setlength{\treeoffsetsix}{\treeoffsetseven}%
\setlength{\treeoffsetseven}{\treeoffseteight}%
\setlength{\treeoffseteight}{\treeoffsetnine}%
\setlength{\treeoffsetnine}{\treeoffsetten}%
\setlength{\treeoffsetten}{\treeoffseteleven}%
\setlength{\treeoffseteleven}{\treeoffsettwelve}%
\setlength{\treeoffsettwelve}{\treeoffsetthirteen}%
\setlength{\treeoffsetthirteen}{\treeoffsetfourteen}%
\setlength{\treeoffsetfourteen}{\treeoffsetfifteen}%
\setlength{\treeoffsetfifteen}{\treeoffsetsixteen}%
\setlength{\treeoffsetsixteen}{\treeoffsetseventeen}%
\setlength{\treeoffsetseventeen}{\treeoffseteighteen}%
\setlength{\treeoffseteighteen}{\treeoffsetnineteen}%
\setlength{\treeoffsetnineteen}{\treeoffsettwenty}%
\setlength{\treeshifttwo}{\treeshiftthree}%
\setlength{\treeshiftthree}{\treeshiftfour}%
\setlength{\treeshiftfour}{\treeshiftfive}%
\setlength{\treeshiftfive}{\treeshiftsix}%
\setlength{\treeshiftsix}{\treeshiftseven}%
\setlength{\treeshiftseven}{\treeshifteight}%
\setlength{\treeshifteight}{\treeshiftnine}%
\setlength{\treeshiftnine}{\treeshiftten}%
\setlength{\treeshiftten}{\treeshifteleven}%
\setlength{\treeshifteleven}{\treeshifttwelve}%
\setlength{\treeshifttwelve}{\treeshiftthirteen}%
\setlength{\treeshiftthirteen}{\treeshiftfourteen}%
\setlength{\treeshiftfourteen}{\treeshiftfifteen}%
\setlength{\treeshiftfifteen}{\treeshiftsixteen}%
\setlength{\treeshiftsixteen}{\treeshiftseventeen}%
\setlength{\treeshiftseventeen}{\treeshifteighteen}%
\setlength{\treeshifteighteen}{\treeshiftnineteen}%
\setlength{\treeshiftnineteen}{\treeshifttwenty}%
\setlength{\treewidthtwo}{\treewidththree}%
\setlength{\treewidththree}{\treewidthfour}%
\setlength{\treewidthfour}{\treewidthfive}%
\setlength{\treewidthfive}{\treewidthsix}%
\setlength{\treewidthsix}{\treewidthseven}%
\setlength{\treewidthseven}{\treewidtheight}%
\setlength{\treewidtheight}{\treewidthnine}%
\setlength{\treewidthnine}{\treewidthten}%
\setlength{\treewidthten}{\treewidtheleven}%
\setlength{\treewidtheleven}{\treewidthtwelve}%
\setlength{\treewidthtwelve}{\treewidththirteen}%
\setlength{\treewidththirteen}{\treewidthfourteen}%
\setlength{\treewidthfourteen}{\treewidthfifteen}%
\setlength{\treewidthfifteen}{\treewidthsixteen}%
\setlength{\treewidthsixteen}{\treewidthseventeen}%
\setlength{\treewidthseventeen}{\treewidtheighteen}%
\setlength{\treewidtheighteen}{\treewidthnineteen}%
\setlength{\treewidthnineteen}{\treewidthtwenty}%
\sbox{\treeboxtwo}{\usebox{\treeboxthree}}%
\sbox{\treeboxthree}{\usebox{\treeboxfour}}%
\sbox{\treeboxfour}{\usebox{\treeboxfive}}%
\sbox{\treeboxfive}{\usebox{\treeboxsix}}%
\sbox{\treeboxsix}{\usebox{\treeboxseven}}%
\sbox{\treeboxseven}{\usebox{\treeboxeight}}%
\sbox{\treeboxeight}{\usebox{\treeboxnine}}%
\sbox{\treeboxnine}{\usebox{\treeboxten}}%
\sbox{\treeboxten}{\usebox{\treeboxeleven}}%
\sbox{\treeboxeleven}{\usebox{\treeboxtwelve}}%
\sbox{\treeboxtwelve}{\usebox{\treeboxthirteen}}%
\sbox{\treeboxthirteen}{\usebox{\treeboxfourteen}}%
\sbox{\treeboxfourteen}{\usebox{\treeboxfifteen}}%
\sbox{\treeboxfifteen}{\usebox{\treeboxsixteen}}%
\sbox{\treeboxsixteen}{\usebox{\treeboxseventeen}}%
\sbox{\treeboxseventeen}{\usebox{\treeboxeighteen}}%
\sbox{\treeboxeighteen}{\usebox{\treeboxnineteen}}%
\sbox{\treeboxnineteen}{\usebox{\treeboxtwenty}}}
\newcommand{\leaf}[1]{%
\ifnum\value{treecount}=20\typeout{QobiTeX warning---Tree stack overflow}\fi%
\addtocounter{treecount}{1}%
\sbox{\treeboxtwenty}{\usebox{\treeboxnineteen}}%
\sbox{\treeboxnineteen}{\usebox{\treeboxeighteen}}%
\sbox{\treeboxeighteen}{\usebox{\treeboxseventeen}}%
\sbox{\treeboxseventeen}{\usebox{\treeboxsixteen}}%
\sbox{\treeboxsixteen}{\usebox{\treeboxfifteen}}%
\sbox{\treeboxfifteen}{\usebox{\treeboxfourteen}}%
\sbox{\treeboxfourteen}{\usebox{\treeboxthirteen}}%
\sbox{\treeboxthirteen}{\usebox{\treeboxtwelve}}%
\sbox{\treeboxtwelve}{\usebox{\treeboxeleven}}%
\sbox{\treeboxeleven}{\usebox{\treeboxten}}%
\sbox{\treeboxten}{\usebox{\treeboxnine}}%
\sbox{\treeboxnine}{\usebox{\treeboxeight}}%
\sbox{\treeboxeight}{\usebox{\treeboxseven}}%
\sbox{\treeboxseven}{\usebox{\treeboxsix}}%
\sbox{\treeboxsix}{\usebox{\treeboxfive}}%
\sbox{\treeboxfive}{\usebox{\treeboxfour}}%
\sbox{\treeboxfour}{\usebox{\treeboxthree}}%
\sbox{\treeboxthree}{\usebox{\treeboxtwo}}%
\sbox{\treeboxtwo}{\usebox{\treeboxone}}%
\sbox{\treeboxone}{\ontop{#1}}%
\sbox{\treeboxone}{\raisebox{-\ht\treeboxone}{\usebox{\treeboxone}}}%
\setlength{\treeoffsettwenty}{\treeoffsetnineteen}%
\setlength{\treeoffsetnineteen}{\treeoffseteighteen}%
\setlength{\treeoffseteighteen}{\treeoffsetseventeen}%
\setlength{\treeoffsetseventeen}{\treeoffsetsixteen}%
\setlength{\treeoffsetsixteen}{\treeoffsetfifteen}%
\setlength{\treeoffsetfifteen}{\treeoffsetfourteen}%
\setlength{\treeoffsetfourteen}{\treeoffsetthirteen}%
\setlength{\treeoffsetthirteen}{\treeoffsettwelve}%
\setlength{\treeoffsettwelve}{\treeoffseteleven}%
\setlength{\treeoffseteleven}{\treeoffsetten}%
\setlength{\treeoffsetten}{\treeoffsetnine}%
\setlength{\treeoffsetnine}{\treeoffseteight}%
\setlength{\treeoffseteight}{\treeoffsetseven}%
\setlength{\treeoffsetseven}{\treeoffsetsix}%
\setlength{\treeoffsetsix}{\treeoffsetfive}%
\setlength{\treeoffsetfive}{\treeoffsetfour}%
\setlength{\treeoffsetfour}{\treeoffsetthree}%
\setlength{\treeoffsetthree}{\treeoffsettwo}%
\setlength{\treeoffsettwo}{\treeoffsetone}%
\setlength{\treeoffsetone}{0.5\wd\treeboxone}%
\setlength{\treeshifttwenty}{\treeshiftnineteen}%
\setlength{\treeshiftnineteen}{\treeshifteighteen}%
\setlength{\treeshifteighteen}{\treeshiftseventeen}%
\setlength{\treeshiftseventeen}{\treeshiftsixteen}%
\setlength{\treeshiftsixteen}{\treeshiftfifteen}%
\setlength{\treeshiftfifteen}{\treeshiftfourteen}%
\setlength{\treeshiftfourteen}{\treeshiftthirteen}%
\setlength{\treeshiftthirteen}{\treeshifttwelve}%
\setlength{\treeshifttwelve}{\treeshifteleven}%
\setlength{\treeshifteleven}{\treeshiftten}%
\setlength{\treeshiftten}{\treeshiftnine}%
\setlength{\treeshiftnine}{\treeshifteight}%
\setlength{\treeshifteight}{\treeshiftseven}%
\setlength{\treeshiftseven}{\treeshiftsix}%
\setlength{\treeshiftsix}{\treeshiftfive}%
\setlength{\treeshiftfive}{\treeshiftfour}%
\setlength{\treeshiftfour}{\treeshiftthree}%
\setlength{\treeshiftthree}{\treeshifttwo}%
\setlength{\treeshifttwo}{\treeshiftone}%
\setlength{\treeshiftone}{0pt}%
\setlength{\treewidthtwenty}{\treewidthnineteen}%
\setlength{\treewidthnineteen}{\treewidtheighteen}%
\setlength{\treewidtheighteen}{\treewidthseventeen}%
\setlength{\treewidthseventeen}{\treewidthsixteen}%
\setlength{\treewidthsixteen}{\treewidthfifteen}%
\setlength{\treewidthfifteen}{\treewidthfourteen}%
\setlength{\treewidthfourteen}{\treewidththirteen}%
\setlength{\treewidththirteen}{\treewidthtwelve}%
\setlength{\treewidthtwelve}{\treewidtheleven}%
\setlength{\treewidtheleven}{\treewidthten}%
\setlength{\treewidthten}{\treewidthnine}%
\setlength{\treewidthnine}{\treewidtheight}%
\setlength{\treewidtheight}{\treewidthseven}%
\setlength{\treewidthseven}{\treewidthsix}%
\setlength{\treewidthsix}{\treewidthfive}%
\setlength{\treewidthfive}{\treewidthfour}%
\setlength{\treewidthfour}{\treewidththree}%
\setlength{\treewidththree}{\treewidthtwo}%
\setlength{\treewidthtwo}{\treewidthone}%
\setlength{\treewidthone}{\wd\treeboxone}}
\newcommand{\branch}[2]{%
\setcounter{branchcount}{#1}%
\ifnum\value{branchcount}=1\sbox{\parentbox}{\ontop{#2}}%
\setlength{\parentoffset}{\treeoffsetone}%
\addtolength{\parentoffset}{-0.5\wd\parentbox}%
\setlength{\daughteroffset}{0in}%
\ifdim\parentoffset<0in%
\setlength{\daughteroffset}{-\parentoffset}%
\setlength{\parentoffset}{0in}\fi%
\setlength{\parentwidth}{\parentoffset}%
\addtolength{\parentwidth}{\wd\parentbox}%
\setlength{\treeoffset}{\daughteroffset}%
\addtolength{\treeoffset}{\treeoffsetone}%
\setlength{\treewidth}{\wd\treeboxone}%
\addtolength{\treewidth}{\daughteroffset}%
\ifdim\treewidth<\parentwidth\setlength{\treewidth}{\parentwidth}\fi%
\sbox{\treebox}{\begin{minipage}{\treewidth}%
\begin{flushleft}%
\hspace*{\parentoffset}\usebox{\parentbox}\\
{\setlength{\unitlength}{2ex}%
\hspace*{\treeoffset}\begin{picture}(0,1)%
\put(0,0){\line(0,1){1}}%
\end{picture}}\\
\vspace{-\baselineskip}
\hspace*{\daughteroffset}%
\raisebox{-\ht\treeboxone}{\usebox{\treeboxone}}%
\end{flushleft}%
\end{minipage}}%
\setlength{\treeoffsetone}{\parentoffset}%
\addtolength{\treeoffsetone}{0.5\wd\parentbox}%
\setlength{\treeshiftone}{0pt}%
\setlength{\treewidthone}{\treewidth}%
\sbox{\treeboxone}{\usebox{\treebox}}%
\else\ifnum\value{branchcount}=2\sbox{\parentbox}{\ontop{#2}}%
\setlength{\branchwidthone}{\treewidthtwo}%
\addtolength{\branchwidthone}{\treeoffsetone}%
\addtolength{\branchwidthone}{-\treeshiftone}%
\addtolength{\branchwidthone}{-\treeoffsettwo}%
\setlength{\branchwidth}{\branchwidthone}%
\setlength{\daughteroffsetone}{\branchwidth}%
\addtolength{\daughteroffsetone}{-\branchwidthone}%
\addtolength{\daughteroffsetone}{-\treeshiftone}%
\setlength{\parentoffset}{-0.5\wd\parentbox}%
\addtolength{\parentoffset}{\treeoffsettwo}%
\addtolength{\parentoffset}{0.5\branchwidth}%
\setlength{\daughteroffset}{0in}%
\ifdim\parentoffset<0in%
\setlength{\daughteroffset}{-\parentoffset}%
\setlength{\parentoffset}{0in}\fi%
\setlength{\parentwidth}{\parentoffset}%
\addtolength{\parentwidth}{\wd\parentbox}%
\setlength{\treeoffset}{\daughteroffset}%
\addtolength{\treeoffset}{\treeoffsettwo}%
\setlength{\treewidth}{\wd\treeboxone}%
\addtolength{\treewidth}{\daughteroffsetone}%
\addtolength{\treewidth}{\treewidthtwo}%
\addtolength{\treewidth}{\daughteroffset}%
\ifdim\treewidth<\parentwidth\setlength{\treewidth}{\parentwidth}\fi%
\sbox{\treebox}{\begin{minipage}{\treewidth}%
\begin{flushleft}%
\hspace*{\parentoffset}\usebox{\parentbox}\\
{\setlength{\unitlength}{0.5\branchwidth}%
\hspace*{\treeoffset}\begin{picture}(2,0.5)%
\put(0,0){\line(2,1){1}}%
\put(2,0){\line(-2,1){1}}%
\end{picture}}\\
\vspace{-\baselineskip}
\hspace*{\daughteroffset}%
\makebox[\treewidthtwo][l]%
{\raisebox{-\ht\treeboxtwo}{\usebox{\treeboxtwo}}}%
\hspace*{\daughteroffsetone}%
\raisebox{-\ht\treeboxone}{\usebox{\treeboxone}}%
\end{flushleft}%
\end{minipage}}%
\setlength{\treeoffsetone}{\parentoffset}%
\addtolength{\treeoffsetone}{0.5\wd\parentbox}%
\setlength{\treeshiftone}{0pt}%
\setlength{\treewidthone}{\treewidth}%
\sbox{\treeboxone}{\usebox{\treebox}}\poptree%
\else\ifnum\value{branchcount}=3\sbox{\parentbox}{\ontop{#2}}%
\setlength{\branchwidthone}{\treewidthtwo}%
\addtolength{\branchwidthone}{\treeoffsetone}%
\addtolength{\branchwidthone}{-\treeshiftone}%
\addtolength{\branchwidthone}{-\treeoffsettwo}%
\setlength{\branchwidthtwo}{\treewidththree}%
\addtolength{\branchwidthtwo}{\treeoffsettwo}%
\addtolength{\branchwidthtwo}{-\treeshifttwo}%
\addtolength{\branchwidthtwo}{-\treeoffsetthree}%
\setlength{\branchwidth}{\branchwidthone}%
\ifdim\branchwidthtwo>\branchwidth%
\setlength{\branchwidth}{\branchwidthtwo}\fi%
\setlength{\daughteroffsetone}{\branchwidth}%
\addtolength{\daughteroffsetone}{-\branchwidthone}%
\addtolength{\daughteroffsetone}{-\treeshiftone}%
\setlength{\daughteroffsettwo}{\branchwidth}%
\addtolength{\daughteroffsettwo}{-\branchwidthtwo}%
\addtolength{\daughteroffsettwo}{-\treeshifttwo}%
\setlength{\parentoffset}{-0.5\wd\parentbox}%
\addtolength{\parentoffset}{\treeoffsetthree}%
\addtolength{\parentoffset}{\branchwidth}%
\setlength{\daughteroffset}{0in}%
\ifdim\parentoffset<0in%
\setlength{\daughteroffset}{-\parentoffset}%
\setlength{\parentoffset}{0in}\fi%
\setlength{\parentwidth}{\parentoffset}%
\addtolength{\parentwidth}{\wd\parentbox}%
\setlength{\treeoffset}{\daughteroffset}%
\addtolength{\treeoffset}{\treeoffsetthree}%
\setlength{\treewidth}{\wd\treeboxone}%
\addtolength{\treewidth}{\daughteroffsetone}%
\addtolength{\treewidth}{\treewidthtwo}%
\addtolength{\treewidth}{\daughteroffsettwo}%
\addtolength{\treewidth}{\treewidththree}%
\addtolength{\treewidth}{\daughteroffset}%
\ifdim\treewidth<\parentwidth\setlength{\treewidth}{\parentwidth}\fi%
\sbox{\treebox}{\begin{minipage}{\treewidth}%
\begin{flushleft}%
\hspace*{\parentoffset}\usebox{\parentbox}\\
{\setlength{\unitlength}{0.5\branchwidth}%
\hspace*{\treeoffset}\begin{picture}(4,1)%
\put(0,0){\line(2,1){2}}%
\put(2,0){\line(0,1){1}}%
\put(4,0){\line(-2,1){2}}%
\end{picture}}\\
\vspace{-\baselineskip}
\hspace*{\daughteroffset}%
\makebox[\treewidththree][l]%
{\raisebox{-\ht\treeboxthree}{\usebox{\treeboxthree}}}%
\hspace*{\daughteroffsettwo}%
\makebox[\treewidthtwo][l]%
{\raisebox{-\ht\treeboxtwo}{\usebox{\treeboxtwo}}}%
\hspace*{\daughteroffsetone}%
\raisebox{-\ht\treeboxone}{\usebox{\treeboxone}}%
\end{flushleft}%
\end{minipage}}%
\setlength{\treeoffsetone}{\parentoffset}%
\addtolength{\treeoffsetone}{0.5\wd\parentbox}%
\setlength{\treeshiftone}{0pt}%
\setlength{\treewidthone}{\treewidth}%
\sbox{\treeboxone}{\usebox{\treebox}}\poptree\poptree%
\else\ifnum\value{branchcount}=4\sbox{\parentbox}{\ontop{#2}}%
\setlength{\branchwidthone}{\treewidthtwo}%
\addtolength{\branchwidthone}{\treeoffsetone}%
\addtolength{\branchwidthone}{-\treeshiftone}%
\addtolength{\branchwidthone}{-\treeoffsettwo}%
\setlength{\branchwidthtwo}{\treewidththree}%
\addtolength{\branchwidthtwo}{\treeoffsettwo}%
\addtolength{\branchwidthtwo}{-\treeshifttwo}%
\addtolength{\branchwidthtwo}{-\treeoffsetthree}%
\setlength{\branchwidththree}{\treewidthfour}%
\addtolength{\branchwidththree}{\treeoffsetthree}%
\addtolength{\branchwidththree}{-\treeshiftthree}%
\addtolength{\branchwidththree}{-\treeoffsetfour}%
\setlength{\branchwidth}{\branchwidthone}%
\ifdim\branchwidthtwo>\branchwidth%
\setlength{\branchwidth}{\branchwidthtwo}\fi%
\ifdim\branchwidththree>\branchwidth%
\setlength{\branchwidth}{\branchwidththree}\fi%
\setlength{\daughteroffsetone}{\branchwidth}%
\addtolength{\daughteroffsetone}{-\branchwidthone}%
\addtolength{\daughteroffsetone}{-\treeshiftone}%
\setlength{\daughteroffsettwo}{\branchwidth}%
\addtolength{\daughteroffsettwo}{-\branchwidthtwo}%
\addtolength{\daughteroffsettwo}{-\treeshifttwo}%
\setlength{\daughteroffsetthree}{\branchwidth}%
\addtolength{\daughteroffsetthree}{-\branchwidththree}%
\addtolength{\daughteroffsetthree}{-\treeshiftthree}%
\setlength{\parentoffset}{-0.5\wd\parentbox}%
\addtolength{\parentoffset}{\treeoffsetfour}%
\addtolength{\parentoffset}{1.5\branchwidth}%
\setlength{\daughteroffset}{0in}%
\ifdim\parentoffset<0in%
\setlength{\daughteroffset}{-\parentoffset}%
\setlength{\parentoffset}{0in}\fi%
\setlength{\parentwidth}{\parentoffset}%
\addtolength{\parentwidth}{\wd\parentbox}%
\setlength{\treeoffset}{\daughteroffset}%
\addtolength{\treeoffset}{\treeoffsetfour}%
\setlength{\treewidth}{\wd\treeboxone}%
\addtolength{\treewidth}{\daughteroffsetone}%
\addtolength{\treewidth}{\treewidthtwo}%
\addtolength{\treewidth}{\daughteroffsettwo}%
\addtolength{\treewidth}{\treewidththree}%
\addtolength{\treewidth}{\daughteroffsetthree}%
\addtolength{\treewidth}{\treewidthfour}%
\addtolength{\treewidth}{\daughteroffset}%
\ifdim\treewidth<\parentwidth\setlength{\treewidth}{\parentwidth}\fi%
\sbox{\treebox}{\begin{minipage}{\treewidth}%
\begin{flushleft}%
\hspace*{\parentoffset}\usebox{\parentbox}\\
{\setlength{\unitlength}{0.5\branchwidth}%
\hspace*{\treeoffset}\begin{picture}(6,1)%
\put(0,0){\line(3,1){3}}%
\put(2,0){\line(1,1){1}}%
\put(4,0){\line(-1,1){1}}%
\put(6,0){\line(-3,1){3}}%
\end{picture}}\\
\vspace{-\baselineskip}
\hspace*{\daughteroffset}%
\makebox[\treewidthfour][l]%
{\raisebox{-\ht\treeboxfour}{\usebox{\treeboxfour}}}%
\hspace*{\daughteroffsetthree}%
\makebox[\treewidththree][l]%
{\raisebox{-\ht\treeboxthree}{\usebox{\treeboxthree}}}%
\hspace*{\daughteroffsettwo}%
\makebox[\treewidthtwo][l]%
{\raisebox{-\ht\treeboxtwo}{\usebox{\treeboxtwo}}}%
\hspace*{\daughteroffsetone}%
\raisebox{-\ht\treeboxone}{\usebox{\treeboxone}}%
\end{flushleft}%
\end{minipage}}%
\setlength{\treeoffsetone}{\parentoffset}%
\addtolength{\treeoffsetone}{0.5\wd\parentbox}%
\setlength{\treeshiftone}{0pt}%
\setlength{\treewidthone}{\treewidth}%
\sbox{\treeboxone}{\usebox{\treebox}}\poptree\poptree\poptree%
\else\ifnum\value{branchcount}=5\sbox{\parentbox}{\ontop{#2}}%
\setlength{\branchwidthone}{\treewidthtwo}%
\addtolength{\branchwidthone}{\treeoffsetone}%
\addtolength{\branchwidthone}{-\treeshiftone}%
\addtolength{\branchwidthone}{-\treeoffsettwo}%
\setlength{\branchwidthtwo}{\treewidththree}%
\addtolength{\branchwidthtwo}{\treeoffsettwo}%
\addtolength{\branchwidthtwo}{-\treeshifttwo}%
\addtolength{\branchwidthtwo}{-\treeoffsetthree}%
\setlength{\branchwidththree}{\treewidthfour}%
\addtolength{\branchwidththree}{\treeoffsetthree}%
\addtolength{\branchwidththree}{-\treeshiftthree}%
\addtolength{\branchwidththree}{-\treeoffsetfour}%
\setlength{\branchwidthfour}{\treewidthfive}%
\addtolength{\branchwidthfour}{\treeoffsetfour}%
\addtolength{\branchwidthfour}{-\treeshiftfour}%
\addtolength{\branchwidthfour}{-\treeoffsetfive}%
\setlength{\branchwidth}{\branchwidthone}%
\ifdim\branchwidthtwo>\branchwidth%
\setlength{\branchwidth}{\branchwidthtwo}\fi%
\ifdim\branchwidththree>\branchwidth%
\setlength{\branchwidth}{\branchwidththree}\fi%
\ifdim\branchwidthfour>\branchwidth%
\setlength{\branchwidth}{\branchwidthfour}\fi%
\setlength{\daughteroffsetone}{\branchwidth}%
\addtolength{\daughteroffsetone}{-\branchwidthone}%
\addtolength{\daughteroffsetone}{-\treeshiftone}%
\setlength{\daughteroffsettwo}{\branchwidth}%
\addtolength{\daughteroffsettwo}{-\branchwidthtwo}%
\addtolength{\daughteroffsettwo}{-\treeshifttwo}%
\setlength{\daughteroffsetthree}{\branchwidth}%
\addtolength{\daughteroffsetthree}{-\branchwidththree}%
\addtolength{\daughteroffsetthree}{-\treeshiftthree}%
\setlength{\daughteroffsetfour}{\branchwidth}%
\addtolength{\daughteroffsetfour}{-\branchwidthfour}%
\addtolength{\daughteroffsetfour}{-\treeshiftfour}%
\setlength{\parentoffset}{-0.5\wd\parentbox}%
\addtolength{\parentoffset}{\treeoffsetfive}%
\addtolength{\parentoffset}{2\branchwidth}%
\setlength{\daughteroffset}{0in}%
\ifdim\parentoffset<0in%
\setlength{\daughteroffset}{-\parentoffset}%
\setlength{\parentoffset}{0in}\fi%
\setlength{\parentwidth}{\parentoffset}%
\addtolength{\parentwidth}{\wd\parentbox}%
\setlength{\treeoffset}{\daughteroffset}%
\addtolength{\treeoffset}{\treeoffsetfive}%
\setlength{\treewidth}{\wd\treeboxone}%
\addtolength{\treewidth}{\daughteroffsetone}%
\addtolength{\treewidth}{\treewidthtwo}%
\addtolength{\treewidth}{\daughteroffsettwo}%
\addtolength{\treewidth}{\treewidththree}%
\addtolength{\treewidth}{\daughteroffsetthree}%
\addtolength{\treewidth}{\treewidthfour}%
\addtolength{\treewidth}{\daughteroffsetfour}%
\addtolength{\treewidth}{\treewidthfive}%
\addtolength{\treewidth}{\daughteroffset}%
\ifdim\treewidth<\parentwidth\setlength{\treewidth}{\parentwidth}\fi%
\sbox{\treebox}{\begin{minipage}{\treewidth}%
\begin{flushleft}%
\hspace*{\parentoffset}\usebox{\parentbox}\\
{\setlength{\unitlength}{0.5\branchwidth}%
\hspace*{\treeoffset}\begin{picture}(8,1)%
\put(0,0){\line(4,1){4}}%
\put(2,0){\line(2,1){2}}%
\put(4,0){\line(0,1){1}}%
\put(6,0){\line(-2,1){2}}%
\put(8,0){\line(-4,1){4}}%
\end{picture}}\\
\vspace{-\baselineskip}
\hspace*{\daughteroffset}%
\makebox[\treewidthfive][l]%
{\raisebox{-\ht\treeboxfour}{\usebox{\treeboxfive}}}%
\hspace*{\daughteroffsetfour}%
\makebox[\treewidthfour][l]%
{\raisebox{-\ht\treeboxfour}{\usebox{\treeboxfour}}}%
\hspace*{\daughteroffsetthree}%
\makebox[\treewidththree][l]%
{\raisebox{-\ht\treeboxthree}{\usebox{\treeboxthree}}}%
\hspace*{\daughteroffsettwo}%
\makebox[\treewidthtwo][l]%
{\raisebox{-\ht\treeboxtwo}{\usebox{\treeboxtwo}}}%
\hspace*{\daughteroffsetone}%
\raisebox{-\ht\treeboxone}{\usebox{\treeboxone}}%
\end{flushleft}%
\end{minipage}}%
\setlength{\treeoffsetone}{\parentoffset}%
\addtolength{\treeoffsetone}{0.5\wd\parentbox}%
\setlength{\treeshiftone}{0pt}%
\setlength{\treewidthone}{\treewidth}%
\sbox{\treeboxone}{\usebox{\treebox}}\poptree\poptree\poptree\poptree%
\else\typeout{QobiTeX warning--- Can't handle #1 branching}\fi\fi\fi\fi\fi}
\newcommand{\tree}{%
\usebox{\treeboxone}
\setlength{\treeoffsetone}{\treeoffsettwo}%
\sbox{\treeboxone}{\usebox{\treeboxtwo}}%
\poptree}

\newcommand{\fs}{\mbox{$/$}}           
\newcommand{\bs}{\mbox{$\backslash$}}  
\newcommand{\us}{\mbox{$\mid$}}        
\newcommand{\exf}{\em}                 
\newcommand{\cgf}{\it}                 
\newcommand{\combf}{\sf}               
\newcommand{\red}{\mbox{$\geq$}}                    
\newcommand{\so}[1]{\mbox{$#1^\prime$\,}}           
\newcommand{\jux}{\mbox{$\cdot$}}                 

\newcommand{\combb}{\mbox{\combf B}}      
\newcommand{\combi}{\mbox{\combf I}}      
\newcommand{\combc}{\mbox{\combf C}}      
\newcommand{\combs}{\mbox{\combf S}}      
\newcommand{\combw}{\mbox{\combf W}}      
\newcommand{\comba}{\mbox{\combf A}}      
\newcommand{\combt}{\mbox{\combf T}}      

\newcommand{\cgfa}{\mbox{\comba$_>$}}   
\newcommand{\cgba}{\mbox{\comba$_<$}}   

\newcommand{\cgfc}{\mbox{\combb$_>$}}   
\newcommand{\cgbc}{\mbox{\combb$_<$}}
\newcommand{\cgfx}{\mbox{\combb$_{\times>}$}}    
\newcommand{\cgbx}{\mbox{\combb$_{\times<}$}}

\newcommand{\cgs}[2]{{\cgf #1\mbox{$_{#2}$}}}
\newcommand{\cgex}[2]                        
  {
  \begin{tabular}[t]{@{}*{#1}{c@{\ }}}       
     #2
  \end{tabular}}
\newcommand{\mc}[2]                     
  {\multicolumn{#1}{c}{#2}}             
\newcommand{\mcl}[1]                    
  {\multicolumn{1}{l}{#1}}              
\newcommand{\cgline}[2]                 
 {\mc{#1}{\hrulefill \raisebox{-.5ex}{\scriptsize #2}}} 
\newcommand{\badline}[2]                
 {\mc{#1}{\hrulefill\raisebox{-.85ex}{***}\hrulefill
\raisebox{-.5ex}{\scriptsize #2}}}      

\newcommand{\cgul}{\cgline{1}{}}        
\newcommand{\cgres}[2]                  
    {\mc{#1}{\cgf #2}}

\newcommand{\acc}{\cgs{NP}{acc}}
\newcommand{\erg}{\cgs{NP}{erg}}

\newcommand{\gnom}{\cgs{NP}{1}}   
\newcommand{\gacc}{\cgs{NP}{2}}

\newcommand{\gdat}{\cgs{NP}{3}}

\newcommand{\ggen}{\cgs{NP}{5}}

\newcommand{\gcativt}{{\cgf S\us\gnom}}                  
\newcommand{\gcattvt}{{\cgf S\us\gnom\us\gacc}}           
\newcommand{\gcatdvt}{{\cgf S\us\gnom\us\gdat\us\gacc}}    
\newcommand{\gcatsubt}{{\cgf S{\fs}(\gcativt)}}  
\newcommand{\gcatobjt}{(\gcativt)\fs(\gcattvt)}           
\newcommand{\gcatobjtb}{(\gcativt)\bs(\gcattvt)}           
\newcommand{\gcatdatt}{(\gcattvt)\fs(\gcatdvt)}           
\newcommand{\gcatgent}{\cgs{NP}{agr}\fs(\cgs{NP}{agr}\bs\ggen)} 

\newcommand{\gcative}{{\cgf S\bs\gnom}}   
\newcommand{\gcattve}{{\cgf (\gcative)\fs\gacc}}   
\newcommand{\gcatsube}{{\cgf S{\fs}(\gcative)}}  
\newcommand{\gcatobjeb}{(\gcative)\bs(\gcattve)}           

\newlength\titlebox
\twocolumn
\def\addcontentsline#1#2#3{}

\def\maketitle{%
  \par%
  \begingroup%
     \def\thefootnote{\fnsymbol{footnote}}%
     \def\@makefnmark{\rlap{$^{\@thefnmark}$\hss}}%
     \long\def\@makefntext##1{%
                  \parindent 1em\noindent%
                  \hbox to 1em{$^{\@thefnmark}$}##1}
     \twocolumn[\@maketitle] \@thanks%
  \endgroup%
  \setcounter{footnote}{0}%
  \let\maketitle\relax\let\@maketitle\relax%
  \gdef\@thanks{}\gdef\@author{}\gdef\@title{}%
  \let\thanks\relax}

\def\@maketitle{%
  \vbox to \titlebox{%
    \hsize\textwidth\linewidth\hsize%
    \vskip 0.125in minus 0.05in%
    \centering{\Large\bf \@title \par}%
    \vskip 0.2in plus 0.1fil minus 0.1in
    {\def\and{\unskip\enspace{\rm and}\enspace}%
     \def\And{\end{tabular}\hss \egroup \hskip 1in plus 2fil 
              \hbox to 0pt\bgroup\hss \begin{tabular}[t]{c}\bf}%
     \def\AND{\end{tabular}\hss\egroup \hfil\hfil\egroup
	      \vskip 0.25in plus 1fil minus 0.125in
	      \hbox to \linewidth\bgroup\large \hfil\hfil
   	      \hbox to 0pt\bgroup\hss \begin{tabular}[t]{c}\bf}
    \hbox to \linewidth \bgroup\large \hfil\hfil
    \hbox to 0pt\bgroup\hss \begin{tabular}[t]{c}\bf\@author 
			    \end{tabular}\hss\egroup
    \hfil\hfil\egroup}
  \vskip 0.3in plus 2fil minus 0.1in
}}

\renewenvironment{abstract}{\section*{\centerline{Abstract}}}{}

\def\@citex[#1]#2{\if@filesw\immediate\write\@auxout{\string\citation{#2}}\fi
  \def\@citea{}\@cite{\@for\@citeb:=#2\do
     {\@citea\def\@citea{; }\@ifundefined
       {b@\@citeb}{{\bf ?}\@warning
        {Citation `\@citeb' on page \thepage \space undefined}}%
 {\csname b@\@citeb\endcsname}}}{#1}}

\let\@internalcite\cite
\def\cite{\def\citename##1{##1, }\@internalcite}
\def\shortcite{\def\citename##1{}\@internalcite}
\def\newcite{\leavevmode\def\citename##1{{##1} (}\@internalciteb}

\def\@citexb[#1]#2{\if@filesw\immediate\write\@auxout{\string\citation{#2}}\fi
  \def\@citea{}\@newcite{\@for\@citeb:=#2\do
    {\@citea\def\@citea{;\penalty\@m\ }\@ifundefined
       {b@\@citeb}{{\bf ?}\@warning
       {Citation `\@citeb' on page \thepage \space undefined}}%
     {\csname b@\@citeb\endcsname}}}{#1}}

\def\@internalciteb{%
  \@ifnextchar [{\@tempswatrue\@citexb}{\@tempswafalse\@citexb[]}}

\def\@newcite#1#2{{#1\if@tempswa, #2\fi)}}

\def\@biblabel#1{}

\def\@cite#1#2{({#1\if@tempswa , #2\fi})}

\def\thebibliography#1{%
  \section*{References}
  \list{}{\setlength{\labelwidth}{0pt}
          \setlength{\leftmargin}{\parindent}
          \setlength{\itemsep}{0.11ex plus 0.11ex}
          \setlength{\parsep}{0ex}
          \setlength{\itemindent}{-\parindent}}
  \def\newblock{\hskip .11em plus .11em minus -.07em}
  \sloppy\clubpenalty4000\widowpenalty4000
  \sfcode`\.=1000\relax}

\def\thesourcebibliography#1{%
  \section*{Sources of Attested Examples}
  \list{}{\setlength{\labelwidth}{0pt}
          \setlength{\leftmargin}{\parindent}
          \setlength{\itemsep}{0.11ex plus 0.11ex}
          \setlength{\parsep}{0ex}
          \setlength{\itemindent}{-\parindent}}
  \def\newblock{\hskip .11em plus .11em minus -.07em}
  \sloppy\clubpenalty4000\widowpenalty4000
  \sfcode`\.=1000\relax}

\def\@lbibitem[#1]#2{\item[]\if@filesw 
      { \def\protect##1{\string ##1\space}\immediate
        \write\@auxout{\string\bibcite{#2}{#1}}\fi\ignorespaces}}

\def\@bibitem#1{\item\if@filesw \immediate\write\@auxout
       {\string\bibcite{#1}{\the\c@enumi}}\fi\ignorespaces}

\def\section{%
    \@startsection{section}{1}{\z@}%
                  {-2.0ex plus -0.5ex minus -0.3ex}%
                  {0.8ex plus 0.3ex minus 0.2ex}%
                  {\large\bf\raggedright}}
\def\subsection{%
    \@startsection{subsection}{2}{\z@}%
                  {-1.4ex plus -0.4ex minus -0.2ex}%
                  {0.6ex plus 0.2ex minus 0.1ex}%
                  {\normalsize\bf\raggedright}}
\def\subsubsection{%
    \@startsection{subsubsection}{3}{\z@}%
                  {-0.8ex plus -0.3ex minus -0.1ex}%
                  {0.4ex plus 0.1ex minus 0.1ex}%
                  {\normalsize\bf\raggedright}}
\def\paragraph{%
    \@startsection{paragraph}{4}{\z@}%
                  {-0.8ex plus -0.3ex minus -0.1ex}%
                  {-1em}%
                  {\normalsize\bf}}
\def\subparagraph{%
    \@startsection{subparagraph}{5}{\parindent}%
                  {0.4ex plus 0.3ex minus 0.1ex}%
                  {-1em}%
                  {\normalsize\bf}}

\labelwidth\leftmargini\advance\labelwidth-\labelsep \labelsep 5pt

\ifcase\@ptsize%
    \renewcommand{\normalsize}{
        \@setsize\normalsize{11.3pt}\xpt\@xpt%
        \abovedisplayskip 10\p@\@plus2\p@\@minus5\p@%
        \abovedisplayshortskip\z@\@plus3\p@%
        \belowdisplayshortskip 4\p@\@plus3\p@\@minus3\p@%
        \belowdisplayskip\abovedisplayskip%
        \let\@listi\@listI}%
 \or%
    \renewcommand{\normalsize}{
        \@setsize\normalsize{12.6pt}\xipt\@xipt%
        \abovedisplayskip11\p@\@plus2\p@\@minus4\p@%
        \abovedisplayshortskip\z@\@plus3\p@%
        \belowdisplayshortskip5\p@\@plus3\p@\@minus2\p@%
        \belowdisplayskip\abovedisplayskip%
        \let\@listi\@listI}%
 \or%
    \renewcommand{\normalsize}{
        \@setsize\normalsize{13pt}\xiipt\@xiipt%
        \abovedisplayskip 11\p@ \@plus3\p@ \@minus5\p@%
        \abovedisplayshortskip \z@ \@plus3\p@%
        \belowdisplayshortskip 5\p@ \@plus3\p@ \@minus2\p@%
        \belowdisplayskip\abovedisplayskip%
        \let\@listi\@listI}%
 \fi    

\ifcase\@ptsize%
    \renewcommand{\small}{
        \@setsize\small{10.5pt}\ixpt\@ixpt%
        \abovedisplayskip 8\p@ \@plus3\p@ \@minus3\p@%
        \abovedisplayshortskip \z@ \@plus2\p@%
        \belowdisplayshortskip 3\p@ \@plus2\p@ \@minus2\p@%
        \belowdisplayskip\abovedisplayskip%
        \def\@listi{\leftmargin\leftmargini%
                    \topsep 3.5\p@ \@plus1.5\p@ \@minus1.5\p@%
                    \parsep 1.5\p@ \@plus\p@ \@minus\p@%
                    \itemsep \parsep}}%
 \or%
    \renewcommand{\small}{
        \@setsize\small{11.3pt}\xpt\@xpt%
        \abovedisplayskip 9\p@ \@plus2\p@ \@minus4\p@%
        \abovedisplayshortskip \z@ \@plus3\p@%
        \belowdisplayshortskip 5\p@ \@plus2.5\p@ \@minus2.5\p@%
        \belowdisplayskip\abovedisplayskip%
        \def\@listi{\leftmargin\leftmargini%
                    \topsep 5\p@ \@plus2\p@ \@minus2\p@%
                    \parsep 2\p@ \@plus2\p@ \@minus\p@%
                    \itemsep \parsep}}%
 \or%
    \renewcommand{\small}{
        \@setsize\small{12pt}\xipt\@xipt%
        \abovedisplayskip 9\p@ \@plus3\p@ \@minus4\p@%
        \abovedisplayshortskip \z@ \@plus3\p@%
        \belowdisplayshortskip 5\p@ \@plus2.5\p@ \@minus2\p@%
        \belowdisplayskip\abovedisplayskip%
        \def\@listi{\leftmargin\leftmargini%
                    \topsep 5.5\p@ \@plus2.5\p@ \@minus2.5\p@%
                    \parsep 4\p@ \@plus1.5\p@ \@minus\p@%
                    \itemsep \parsep}}%
 \fi

\ifcase\@ptsize
    \renewcommand{\footnotesize}{
        \@setsize\footnotesize{9.3pt}\viiipt\@viiipt%
        \abovedisplayskip 5\p@ \@plus2\p@ \@minus3\p@%
        \abovedisplayshortskip \z@ \@plus\p@%
        \belowdisplayshortskip 2.5\p@\@plus\p@\@minus2\p@%
        \belowdisplayskip\abovedisplayskip%
        \def\@listi{\leftmargin\leftmargini%
                    \topsep 2.5\p@ \@plus\p@ \@minus\p@%
                    \parsep 1.5\p@ \@plus\p@ \@minus\p@%
                    \itemsep \parsep}}%
 \or%
    \renewcommand{\footnotesize}{
        \@setsize\footnotesize{10.3pt}\ixpt\@ixpt%
        \abovedisplayskip 7\p@ \@plus2\p@ \@minus4\p@%
        \abovedisplayshortskip \z@ \@plus\p@%
        \belowdisplayshortskip 3\p@ \@plus2\p@ \@minus2\p@%
        \belowdisplayskip\abovedisplayskip%
        \def\@listi{\leftmargin\leftmargini%
                    \topsep 3\p@ \@plus2\p@ \@minus2\p@%
                    \parsep 2\p@ \@plus\p@ \@minus\p@%
                    \itemsep \parsep}}%
 \or%
    \renewcommand{\footnotesize}{
        \@setsize\footnotesize{11pt}\xpt\@xpt%
        \abovedisplayskip 9\p@ \@plus2\p@ \@minus4\p@%
        \abovedisplayshortskip \z@ \@plus3\p@%
        \belowdisplayshortskip 5\p@ \@plus3\p@ \@minus3\p@%
        \belowdisplayskip\abovedisplayskip%
        \def\@listi{\leftmargin\leftmargini%
                    \topsep 4.5\p@ \@plus2\p@ \@minus2\p@%
                    \parsep 3\p@ \@plus\p@ \@minus\p@%
                    \itemsep \parsep}}%
 \fi

\ifcase\@ptsize%
    \renewcommand{\large}{\@setsize\large{13pt}\xiipt\@xiipt}
 \or%
    \renewcommand{\large}{\@setsize\large{13pt}\xiipt\@xiipt}
 \or%
    \renewcommand{\large}{\@setsize\large{16pt}\xivpt\@xivpt}
 \fi

\ifcase\@ptsize%
    \renewcommand{\Large}{\@setsize\Large{16pt}\xivpt\@xivpt}
 \or%
    \renewcommand{\Large}{\@setsize\Large{16pt}\xivpt\@xivpt}
 \or%
    \renewcommand{\Large}{\@setsize\Large{16pt}\xivpt\@xivpt}
 \fi

\ifcase\@ptsize%
 \or%
 \or%
 \fi

\ifcase\@ptsize%
    \def\@listi{\leftmargin\leftmargini
                \topsep  6\p@ \@plus2\p@ \@minus2\p@%
                \parsep  2\p@ \@plus0.5\p@ \@minus\p@%
                \itemsep 2.5\p@ \@plus\p@ \@minus0.5\p@}%
 \or%
    \def\@listi{\leftmargin\leftmargini
                \topsep  8\p@ \@plus2\p@ \@minus2\p@%
                \parsep  3\p@ \@plus1.5\p@ \@minus\p@%
                \itemsep 3\p@ \@plus1.5\p@ \@minus\p@}%
 \or%
    \def\@listi{\leftmargin\leftmargini
                \topsep  9\p@ \@plus3\p@   \@minus4\p@%
                \parsep  4\p@  \@plus2\p@ \@minus\p@%
                \itemsep 4\p@  \@plus2\p@ \@minus\p@}%
 \fi
\let\@listI\@listi

\newcommand{\exm}[1]{{\it #1}\/}
\newcommand{\app}{\mbox{$\cdot$}}
 \usepackage{times}
 \usepackage{lingmacros}
 \usepackage{tree-dvips}
 \usepackage{avm}
\title{Deriving the Predicate-Argument Structure for a Free Word Order
Language
\thanks{Thanks to Mark Steedman for discussion and material, and
to the anonymous reviewer of an extended version whose comments led
to significant revisions. This research is supported by
TUBITAK (EEEAG-90) and NATO Science Division (TU-LANGUAGE).}}
 \author{Cem Bozsahin\\Department of Computer Engineering\\
Middle East Technical University\\ 06531 Ankara, Turkey\\
{\tt bozsahin@ceng.metu.edu.tr}}

\begin{document}
 \maketitle
 \begin{abstract}
In relatively free word order languages, grammatical functions 
are intricately related to case marking. 
Assuming an ordered representation 
of the predicate-argument structure, this work proposes a Combinatory
Categorial Grammar formulation of relating surface case cues to categories
and types
for correctly placing the arguments in the predicate-argument structure. 
This is achieved by assigning case markers GF-encoding categories. 
Unlike other CG 
formulations, type shifting does not proliferate or cause spurious ambiguity. 
Categories of all argument-encoding grammatical functions follow from the
same principle of category assignment. 
Normal order evaluation of the combinatory 
form reveals the predicate-argument structure. The application of the method 
to Turkish is shown.
\end{abstract}

\section{Introduction}\label{sec:introduction}
Recent theorizing in linguistics brought forth a level of representation
called the Predicate-Argument Structure (PAS). PAS acts as the interface 
between lexical semantics and d-structure in GB \cite{grimshaw90}, 
functional structure in LFG \cite{alsina96}, and complement structure
in HPSG \cite{wechsler95}. PAS is the sole level of representation in
Combinatory Categorial Grammar (CCG) \cite{steedman96}. All formulations
assume a prominence-based structured representation for PAS, although they
differ in the terms used for defining prominence.
For instance, Grimshaw \shortcite{grimshaw90} defines the thematic hierarchy as:

Agent $>$ Experiencer $>$ Goal / Location / Source 

$>$ Theme

\noindent whereas LFG accounts make use of the following 
\cite{bresnankanerva89}:

Agent $>$ Beneficiary $>$ Goal / Experiencer $>$ Inst

$>$ Patient / Theme $>$ Locative.

As an illustration, the predicate-argument structures of the agentive verb
\exm{murder} and the psychological verb \exm{fear} are
\cite[p.8]{grimshaw90}:

\begin{tabular}{lll} 
\exm{murder} & (x & (y))\\
& Agent & Theme \\
\exm{fear} & (x & (y))\\
& Exp & Theme
\end{tabular}

To abstract away from language-particular case systems and mapping of thematic
roles to grammatical functions, I assume the Applicative Hierarchy of
Shaumyan \shortcite{shaumyan87} for the definition of prominence:

Primary Term $>$ Secondary Term $>$ 

Tertiary Term $>$ Oblique Term.

\noindent Primacy of a term over another is defined by the former having 
a wider range of syntactic features than the latter. In an accusative
language, subjects are less marked (hence primary) than objects; all verbs 
take
subjects but only transitive verbs take objects. Terms (=arguments) can be
denoted by the {\em genotype}\/ indices on NPs, 
such as \gnom, \gacc\
for primary and secondary terms.\footnote{Shaumyan uses
$T^1, T^2$, but we prefer \gnom, \gacc\ for
easier exposition in later formulations.} 
An \gacc\ would be a direct object (\acc) in an accusative language, or
an ergative-marked NP (\erg) in an ergative language. This level of description
also simplifies the formulation of grammatical function changing; the
primary term of a passivized predicate $(\mbox{\sc pass\ } p)$ 
is the secondary term
of the active $p$. I follow Shaumyan and Steedman 
\shortcite{steedman96} also in the ordered representation of the PAS
(\ref{ex:pas}). The reader is referred to \cite{shaumyan87} for
linguistic justification of this ordering.

\enumsentence{\label{ex:pas}
Pred$\ldots<$Sec. Term$><$Primary Term$>$
}

Given this representation, the surface order of constituents is often in
conflict with the order in the PAS. For instance, English as a configurational
SVO language has the mapping:

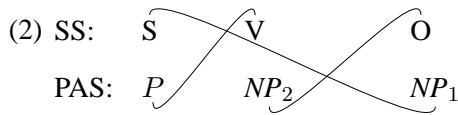
\enumsentence{\label{ex:svo}
\begin{avm} 
\avml SS: & \!{a}{S}\hspace*{2em} & \!{c}{V}\hspace*{4em} & \!{d}{O}\\[2ex]
PAS: & \!{e}{$P$} & \!{f}{\gacc} & \!{g}{\gnom} \avmr
\end{avm}
\nodecurve[t]{a}[b]{g}{15pt}
\nodecurve[t]{c}[b]{e}{15pt}
\nodecurve[t]{d}[b]{f}{15pt}
}

However, in a non-configurational language, permutations 
of word order are possible, and grammatical functions are often indicated
not by configurations but by case marking. For instance, in Turkish,
all six permutations of the basic SOV order are possible, and Japanese 
allows two verb-final permutations of underlying SOV. 
The relationship between case marking and scrambling is crucial
in languages with flexible word order.
A computational solution to the problem must rely on some
principles of parsimony for representing categories and types of
arguments and predicates, and efficiency of processing.

In a categorial formulation, grammatical functions of preverbal and
postverbal NPs in (\ref{ex:svo}) can be made explicit by 
type shifting\footnote{aka. type raising, lifting, or type change}
the subject
to \gcatsube\ and the object to \gcatobjeb.
These categories follow from the order-preserving type shifting
scheme \cite{dowty88}:

\enumsentence{\label{rul:ts}
{\cgf NP $\Rightarrow$ T\fs(T\bs NP)}\/ or {\cgf T\bs(T\fs NP)}
}

To resolve the
opposition between surface order and the PAS in a free word order
language, 
one can let the type shifted categories of terms proliferate, or
reformulate CCG in such  a way that arguments of the verbs are sets, rather
than lists whose arguments are made available one at a time. The former
alternative makes the spurious ambiguity problem 
of CG parsing \cite{karttunen89} 
even more severe. Multi-set CCG \cite{hoffman95} is an example
of the set-oriented approach. It is known to be computationally tractable but
less efficient than the polynomial time CCG algorithm of
Vijay-Shanker and Weir \shortcite{vijay-shankerweir93}. I try to show
in this paper that the traditional curried notation of CG with type shifting
can be maintained to account for Surface Form$\leftrightarrow$PAS mapping
without leading to proliferation of argument categories or to
spurious ambiguity.

Categorial framework is particularly suited for this mapping due to its
lexicalism. Grammatical functions of the nouns in the lexicon are assigned
by case markers, which are also in the lexicon. Thus, grammatical function
marking follows naturally from the general CCG schema comprising
rules of application (\comba) and composition (\combb). 
The functor-argument distinction
in CG helps to model prominence relations without extra levels of
representation. CCG schema (Steedman \shortcite{steedman88,steedman90}) is 
summarized
in (\ref{rul:ccg}). Combinator notation is preferred here because they are 
the formal
primitives operating on the PAS (cf. \cite{curryfeys58}
for Combinatory Logic). Application is the only primitive of the combinatory
system; it is indicated by juxtaposition in the examples and denoted by
\app\ in the normal order evaluator (\S\ref{sec:reducer}). \combb\ has
the reduction rule $\combb fga \red f(ga)$.

\enumsentence{\label{rul:ccg}
\begin{tabular}[t]{llll}
\cgf X{\fs}Y$\colon f$ & \cgf Y$\colon a$ & 
   $\Rightarrow_{\mbox{\tiny\cgfa}}$ &  \cgf X$\colon fa$ \\[1.2ex] 
 \cgf Y$\colon a$ & \cgf X\bs Y$\colon f$ & 
   $\Rightarrow_{\mbox{\tiny\cgba}}$&   \cgf X$\colon fa$ \\[1.2ex] 
 \cgf X{\fs}Y$\colon f$ & \cgf Y{\fs}Z$\colon g$ & 
   $\Rightarrow_{\mbox{\tiny\cgfc}}$ &  \cgf X{\fs}Z$\colon \combb fg$ \\[1.2ex]
 \cgf Y\bs Z$\colon g$ & \cgf X\bs Y$\colon f$ & 
   $\Rightarrow_{\mbox{\tiny\cgbc}}$&   \cgf X\bs Z$\colon \combb fg$ \\[1.2ex]
 \cgf X{\fs}Y$\colon f$ & \cgf Y\bs Z$\colon g$ &
 $\Rightarrow_{\mbox{\tiny\cgfx}}$&   \cgf X\bs Z$\colon \combb fg$ \\[1.2ex]
 \cgf Y{\fs}Z$\colon g$ & \cgf X\bs Y$\colon f$ & 
   $\Rightarrow_{\mbox{\tiny\cgbx}}$&   \cgf X{\fs}Z$\colon \combb fg$ \\[1.2ex]
\end{tabular}
}

\section{Grammatical Functions, Type Shifting, and Composition}\label{sec:gfts}
In order to derive all permutations of a ditransitive construction in Turkish
using (\ref{rul:ts}), the
dative-marked indirect object (\gdat) must be type shifted in 48 ($4!2$)
different ways so that coordination with the left-adjacent and the 
right-adjacent constituent is possible. This is due to the fact that
the result category {\cgf T}\/ is always a conjoinable type, and the
argument category {\cgf T\fs\gdat} (and {\cgf T\bs\gdat}) must be allowed 
to compose with the result category of the adjacent functor. However,
categories of arguments can be made more informative about grammatical 
functions and word order. The basic principle is as follows: The category
assigned for argument $n$ must contain all and only the term
information about \cgs{NP}{i} for all $i\leq n$. An \gacc\ type
must contain in its category word order information about \gnom\ and \gacc\ 
but not \gdat.
This can be generalized as in (\ref{rul:tgf}): 

\eenumsentence{\label{rul:tgf} 
\item [] Category assignment for argument $n$:

\item[] \[ {\cal C}(n) = \left\{ \begin{array}{l} 
                         T_r\fs T_a \mbox{\rm\ \  or\ \ }
 T_r\bs T_a\\[1.2ex]
 \cgs{NP}{n}
\end{array} \right. \]

\item[] $T_a=$ Lexical category of an \cgs{NP}{n}-governing element (e.g., a 
verb) in the language whose highest genotype argument is \cgs{NP}{n}.

\item[] $T_r=$ The category obtained from $T_a$ by removing \cgs{NP}{n}.
}

Case markers in Turkish are suffixes attached to noun groups.\footnote{As 
suggested in
\cite{bozsahingocmen95}, morphological and syntactic composition
can be distinguished by associating several attachment calculi
with functors and arguments (e.g., affixation, concatenation,
clitics, etc.)} 
The types of case markers in the lexicon can be defined as:

\eenumsentence{\label{rul:lextype} 
\item [] Lexical type assignment for the case marker
({\em -case}) encoding argument $n$:

\item[] {\em -case}\/{\tt :=}
${\cal C}(n)\colon{\cal T}({\cal C}(n))x\bs N\colon x$
}

where ${\cal T}(C)$ denotes the semantic type for category $C$:

\eenumsentence{\label{rul:sem}
\item[a.] ${\cal T}(\cgs{NP}{n})=\combi$ (lower type for \cgs{NP}{n})
\item[b.] ${\cal T}(C)=\combt$ (if $C$ is a type shifted category as in
(\ref{rul:ts}))
\item[c.] ${\cal T}(C)=\combb\combb\combt$ (if $C$ is a type shifted and
composed category)
}

(\ref{rul:tgf}) and (\ref{rul:lextype}) are schemas that yield three lexical
categories per {\em -case}: one for lower type, and two for higher
types which differ only in the directionality of the main function due
to (\ref{rul:tgf}). For instance, for the accusative case suffix encoding
\gacc, we have:

\noindent\begin{tabular}{rl}
-ACC \tt := &$\gacc\colon\combi x\bs N\colon x$\\
\tt := &$(\gcatobjt)\colon\combt x\bs N\colon x$\\
\tt := &$(\gcatobjtb)\colon\combt x\bs N\colon x$\\
\end{tabular}

Type shifting alone is too constraining if the verbs
take their arguments in an order different from the Applicative Hierarchy
(\S\ref{sec:introduction}). For instance, the category of Turkish
ditransitives is \gcatdvt. Thus the verb has the wrapping semantics
$\combc\so{v}$ where \combc\ is the permutator with the
reduction rule $\combc fga\red fag$. Type shifting an \gdat\ yields 
{\cgf (\gcattvt)\fs(S\us\cgs{NP}{1}\us\cgs{NP}{2}\us\cgs{NP}{3})} 
in which the argument category is not lexically licensed.
(\ref{rul:tgf}) is order-preserving in a language-particular way; the
result category always corresponds to a lexical category in the language
if the argument category does too. 

For arguments requiring a non-canonical order, we need type shifting and
composition (hence the third clause in (\ref{rul:sem})):

\noindent\begin{tabular}{lll}
\cgf \gdat$\colon x$ & $\stackrel{\combt}{\Rightarrow}$ & 
\cgf (\gcativt)\fs(S\us\cgs{NP}{1}\us\cgs{NP}{3})$\colon\combt x$
 $\stackrel{\combb}{\Rightarrow}$  \\
\mc{3}{
\cgf \gcatdatt$\colon\combb(\combt x)=\combb\combb\combt x$}
\end{tabular}

Once syntactic category of the argument is fixed, its semantics is uniquely
determined by (\ref{rul:sem}).

The combinatory primitives operating on the PAS
are 
\combi\ (\ref{rul:sem}a), \combt\ (\ref{rul:sem}b--c), and 
\combb\ (\ref{rul:sem}c). \combt\ has the
reduction rule $\combt af\red fa$, and $\combi f\red f$. The use of
\combt\ or \combb\ signifies 
that the term's category is a functor; its correct place
in the PAS is yet to be determined. \combi\ indicates that the term is
in the right place in the partially derived PAS.

According to (\ref{rul:tgf}), there is a unique result-argument combination
for a higher type \gdat, compared to 24 using (\ref{rul:ts}). 
(\ref{rul:tgf}) differs from (\ref{rul:ts}) in another significant aspect:
$T_r$ and $T_a$ may contain directionally underspecified categories
if licensed by the lexicon. 
Directional underspecification is needed when arguments
of a verb can scramble to either side of the verb. It is necessary
in Turkish and Warlpiri but not in Japanese or Korean.
The neutral slash \us\ is a lexical operator; it is instantiated to
either \bs\ or \fs\ during parsing. A crucial use of underspecification
is shown in (\ref{ex:underspec}). SV composition could not follow through
if the verbs had backward-looking categories; composition of the
type shifted subject and the verb in this case would only yield a 
backward-looking {\cgf S\bs\gacc}\/ by the schema (\ref{rul:ccg}).

{\scriptsize
\enumsentence{\label{ex:underspec}
\cgex{5}{\exf Adam & \exf kurmu\c{s} &\exf  ama &\exf  \c{c}ocuk toplad\i 
&\exf  masa-y\i\\
man.NOM & set & but & child.NOM gather & table-ACC\\
\cgul & \cgul & & \cgline{1}{\cgfc} & \cgul\\
\gcatsubt & \gcattvt & &\cgf S\fs\gacc & \gacc\\
\cgline{2}{\cgfc}  \\
\cgres{2}{S\fs\gacc}  \\
\cgline{4}{$\wedge$}\\ \cgres{4}{S\fs\gacc}\\
\cgline{5}{\cgfa}\\ \cgres{5}{S}\\
\mc{5}{'The man had set the table but the child is cleaning it.'}
}}}

The schema in (\ref{rul:tgf}) makes the arguments available in higher types,
and allows lower (\cgs{NP}{n}) types only if higher types fail (as in
\gacc\ in (\ref{ex:underspec})). There are two reasons for this:
Higher types carry more information about surface order of the language, 
and they are sufficient to cover bounded phenomena. 
\S\ref{sec:wo}
shows how higher types correctly derive the PAS in various word orders.
Lower types are
indispensable for unbounded constructions such as relativization and
coordination. 
The choice is due to a concern for economy. 
If lower types were allowed freely, they would yield the correct PAS as well:

\enumsentence{\label{ex:okpas}
\cgex{4}{S & IO & DO & V\\ \cgul & \cgul & \cgul & \cgul\\
\gnom$\colon\combi\so{s}$ & \gdat$\colon\combi\so{i}$&
\gacc$\colon\combi\so{o}$ & \cgf DV$\colon\combc\so{v}$\\
& & \cgline{2}{\cgba}\\ & & \cgres{2}{S\us\gnom\us\gdat
$\colon(\combc\so{v})(\combi\so{o})$}\\
& \cgline{3}{\cgba}\\ & \cgres{3}{S\us\gnom 
$\colon(\combc\so{v})(\combi\so{o})(\combi\so{i})$}\\
\cgline{4}{\cgba}\\ \cgres{4}{S
$\colon(\combc\so{v})(\combi\so{o})(\combi\so{i})(\combi\so{s})\red
 \so{v}\so{i}\so{o}\so{s}$}
}}

In parsing this is achieved as follows: An \cgs{NP}{k} can only be the 
argument in a rule of application, and schema (\ref{rul:tgf}) is the only
way to obtain \cgs{NP}{k} from a noun group. Thus it suffices to check
in the application rules that if the argument category is \cgs{NP}{k}, then
the functor's result category (e.g., {\cgf X}\/ in {\cgf X\fs Y}) 
has none of the terms with genotype indices lower than $k$.
\gacc\ in (\ref{ex:underspec}) is licensed because the adjacent functor
is {\cgf S\fs\gacc}. \gacc\ in (\ref{ex:okpas}) is not licensed 
because the adjacent functor has \gnom.

For noun-governed grammatical functions such as the genitive (\ggen), 
(\ref{rul:tgf}) licenses result categories that are underspecified
with respect to the genotype index. This is indeed necessary because
the resulting NP can be further inflected on case and assume a genotype index.
For Turkish, the type shifted
category is {\cgf ${\cal C}(5)=$\gcatgent}. 
Hence the genitive suffix bears the category {\cgf ${\cal C}(5)$\bs N}. 
Agreement features enforce the possessor-possessed agreement on person and
number via unification (as in UCG \cite{calderkleinzeevat88}):

\noindent{\footnotesize
\cgex{4}{\exf kalem &\exf -in & \exf uc &\exf -u\\
pencil & -GEN.3s & tip & -POSS.3s\\ \cgul & \cgul & \cgul & \cgul\\
\cgf N$\colon\so{p}$ &\cgf ${\cal C}(5)$\bs N$\colon\combt$ &
\cgf N$\colon\so{t}$ & \cgf (\cgs{NP}{agr}\bs\ggen)\bs N$\colon poss$\\
\cgline{2}{\cgba} & \cgline{2}{\cgba}\\
\cgres{2}{\gcatgent$\colon\combt\so{p}$} & 
\cgres{2}{\cgs{NP}{agr}\bs\ggen$\colon poss\,\so{t}$}\\
\cgline{4}{\cgfa}\\ \cgres{4}{\cgs{NP}{agr}
$\colon\combt\so{p}(poss\,\so{t}) \red (poss\,\so{t})\so{p}$}\\
\mc{4}{'The tip of the pencil'}}}

\section{Word Order and Scrambling}\label{sec:wo}

Due to space limitations, the following abbreviated categories are employed in 
derivations:
\begin{tabular}{lll}
\cgf IV & = & \gcativt\\
\cgf TV & = & \gcattvt\\
\cgf DV & = & \gcatdvt
\end{tabular}

The categories licensed by (\ref{rul:tgf}) can then be written as
{\cgf IV\fs TV}\/ and {\cgf IV\bs TV}\/ for \gacc, 
{\cgf TV\fs DV}\/ and {\cgf TV\bs DV} for \gdat, etc.
(\ref{ex:vfinal}a--b) show the verb-final variations in the word order.
The bracketings in the PAS and juxtaposition are left-associative; $(fa)b$ is 
same as $fab$.

\eenumsentence{\label{ex:vfinal}
\item[a.]\cgex{3}{\exf Mehmet & \exf kitab-\i &\exf oku-du\\
M.NOM & book-ACC & read-PAST\\ \cgul & \cgul & \cgul\\
\cgf S\fs IV$\colon\combt\so{m}$ &
\cgf IV\fs TV$\colon\combt\so{b}$ & \cgf TV$\colon\so{r}$\\
& \cgline{2}{\cgfa}\\ & \cgres{2}{IV$\colon\combt\so{b}\so{r}$}\\
\cgline{3}{\cgfa}\\ \cgres{3}{S$\colon\combt\so{m}(\combt\so{b}\so{r})
\red \so{r}\so{b}\so{m}$}\\
\mc{3}{'Mehmet read the book.'} }
\item[b.]\cgex{3}{\exf kitab-\i & \exf  Mehmet &\exf oku-du\\
\cgul & \cgul & \cgul\\
\cgf IV\fs TV$\colon\combt\so{b}$ &
\cgf S\bs IV$\colon\combt\so{m}$ &
 \cgf TV$\colon\so{r}$\\
\cgline{2}{\cgbx}\\ \cgres{2}{S\fs TV
$\colon\combb(\combt\so{m})(\combt\so{b})$}\\
\cgline{3}{\cgfa}\\ \cgres{3}{S
$\colon\combb(\combt\so{m})(\combt\so{b})\so{r}
\red \so{r}\so{b}\so{m}$}\\
 }
}

(\ref{ex:vfinal}a) exhibits spurious ambiguity. Forward composition of
{\cgf S\fs IV}\/ and {\cgf IV\fs TV} is possible, yielding exactly the
same PAS. This problem is resolved by grammar rewriting in the sense
proposed by Eisner\footnote{Eisner \shortcite[p.81]{eisner96} in fact 
suggested that
the labeling system can be implemented in the grammar by templates, or
in the processor by labeling the chart entries.} 
\shortcite{eisner96}. Grammar rewriting can be done using 
predictive combinators \cite{wittenburg87}, but they cannot handle
crossing compositions that are essential to our method. Other normal form
parsers, e.g. that of Hepple and Morrill \shortcite{hepplemorrill89}, 
have the same problem.
All grammar rules in (\ref{rul:ccg}) in fact check the labels of the
constituent categories, which show how the category is derived. 
The labels are as in \cite{eisner96}. {\tt -FC}: Output of forward composition, 
of which 
forward crossing composition is a special case. {\tt -BC}: Output of
backward composition, of which backward crossing composition is a special
case. {\tt -OT}: Lexical or type shifted category. The goal is to block
e.g., {\cgf X\fs Y}{\tt -FC}\ {\cgf Y\fs Z}{\tt -\{FC,BC,OT\}} 
$\Rightarrow_{\mbox{\cgfc}}$ {\cgf X\fs Z}\/ and {\cgf X\fs Y}{\tt -FC}\ 
{\cgf Y}{\tt -\{FC,BC,OT\}} $\Rightarrow_{\mbox{\cgfa}}$ 
{\cgf X}\/ in (\ref{ex:vfinal}a). {\cgf S\fs TV} composition would have the
label {\tt -FC}, which cannot be an input to forward application. 
In (\ref{ex:vfinal}b), the backward composition follows through since 
it has the category-label {\cgf S\fs TV}{\tt -BC}, which the forward 
application rule does not block. We use Eisner's method to rewrite 
all rules in (\ref{rul:ccg}). 

(\ref{ex:nonfinal}a--b) show the normal form parses for post-verbal
scrambling, and (\ref{ex:nonfinal}c--d) for verb-medial cases. 

\eenumsentence{\label{ex:nonfinal}
\item[a.]\cgex{3}{\exf oku-du &\exf Mehmet & \exf kitab-\i \\
read-PAST & M.NOM & book-ACC\\ \cgul & \cgul & \cgul\\
\cgf TV$\colon\so{r}$&
\cgf S\fs IV$\colon\combt\so{m}$ &
\cgf IV\bs TV$\colon\combt\so{b}$ \\
& \cgline{2}{\cgfx}\\ & \cgres{2}{S\bs TV
$\colon\combb(\combt\so{m})(\combt\so{b})$}\\
\cgline{3}{\cgba}\\ 
\cgres{3}{S
$\colon\combb(\combt\so{m})(\combt\so{b})\so{r} \red
\so{r}\so{b}\so{m}$}\\
\mc{3}{'Mehmet read the book.'} }

\item[b.]\cgex{3}{\exf oku-du &
\exf kitab-\i &
\exf Mehmet \\ 
\cgul & \cgul & \cgul\\
\cgf TV$\colon\so{r}$&
\cgf IV\bs TV$\colon\combt\so{b}$ &
\cgf S\bs IV$\colon\combt\so{m}$ \\
\cgline{2}{\cgba}\\ \cgres{2}{IV
$\colon\combt\so{b}\so{r}$}\\
\cgline{3}{\cgba}\\ 
\cgres{3}{S
$\colon\combt\so{m}(\combt\so{b}\so{r}) \red
\so{r}\so{b}\so{m}$}}

\item[c.]\cgex{3}{
\exf kitab-\i &
\exf oku-du &
\exf Mehmet \\ 
\cgul & \cgul & \cgul\\
\cgf IV\fs TV$\colon\combt\so{b}$ &
\cgf TV$\colon\so{r}$&
\cgf S\bs IV$\colon\combt\so{m}$ \\
\cgline{2}{\cgfa}\\ \cgres{2}{IV
$\colon\combt\so{b}\so{r}$}\\
\cgline{3}{\cgba}\\ 
\cgres{3}{S
$\colon\combt\so{m}(\combt\so{b}\so{r}) \red
\so{r}\so{b}\so{m}$}}

\item[d.]\cgex{3}{
\exf Mehmet &
\exf oku-du &
\exf kitab-\i \\
\cgul & \cgul & \cgul\\
\cgf S\fs IV$\colon\combt\so{m}$ &
\cgf TV$\colon\so{r}$&
\cgf IV\bs TV$\colon\combt\so{b}$ \\
& \cgline{2}{\cgba}\\ & \cgres{2}{IV
$\colon\combt\so{b}\so{r}$}\\
\cgline{3}{\cgfa}\\ 
\cgres{3}{S
$\colon\combt\so{m}(\combt\so{b}\so{r}) \red
\so{r}\so{b}\so{m}$}}
}

Controlled lexical redundancy of higher types,
e.g., having both (and only) {\cgf IV\fs TV}\/ and {\cgf IV\bs TV}\/ licensed
by the lexicon for an \gacc, does not lead to alternative derivations
in (\ref{ex:vfinal}--\ref{ex:nonfinal}). 
Assume that
{\cgf A\fs B\ \ B\bs C},
where {\cgf A\fs B}\/
and {\cgf B\bs C}\/ are categories produced by (\ref{rul:tgf}),
gives a successful parse using the output
{\cgf A\bs C}. 
{\cgf A\bs B\ \ B\bs C}\/ and {\cgf A\bs B\ \ B\fs C}\/ are
not composable types according to (\ref{rul:ccg}). The other possible
configuration, {\cgf A\fs B\ \ B\fs C}, yields an
{\cgf A\fs C}\/ which looks for {\cgf C}\/ in the other direction.
Multiple derivations appear to be possible if there is an order-changing 
composition
over {\cgf C}, such as {\cgf C\fs C} (e.g., a VP modifier
{\cgf IV\fs IV}). (\ref{ex:abc}) shows two possible
configurations with a {\cgf C}\/ on the right. (\ref{ex:abc}b) is blocked
by label check because {\cgf A\fs C{\tt-FC}\ \ C} 
$\Rightarrow_{\mbox{\cgfa}}$ {\cgf A}\/ is not licensed by the grammar.
If {\cgf C}\/ were to the left, only (\ref{ex:abc}a) would succeed.
Similar reasoning can be used to show the uniqueness of derivation 
in other patterns of directions.

\eenumsentence{\label{ex:abc}
\item[a.]\cgex{4}{\cgf C\fs C &\cgf A\fs B & \cgf B\bs C & \cgf C\\
& \cgline{2}{\cgfx}\\ & \cgres{2}{A\bs C{\tt-FC}}\\
\cgline{3}{\cgbx}\\ \cgres{2}{A\fs C{\tt-BC}}\\
\cgline{4}{\cgfa}\\ \cgres{4}{A{\tt-OT}}
}
\item[b.]\cgex{4}{\cgf C\fs C &\cgf A\fs B & \cgf B\fs C & \cgf C\\
& \cgline{2}{\cgfc}\\ & \cgres{2}{A\fs C{\tt-FC}}\\
& \badline{3}{\cgfa}
}}

Constrained type shifting avoids the problem with freely available
categories in Eisner's normal form parsing scheme. However,
some surface characteristics of the language, such as lack of
case marking in certain constructions, puts the burden of type shifting 
on the processor \cite{bozsahin97}. 
Lower type arguments such as \gacc\ pose a different kind of ambiguity
problem. Although they are required in unbounded constructions,
they may yield alternative derivations of local scrambling cases in a 
labelled CCG. For instance, when \gacc\ is peripheral in a ditransitive
construction and the verb can form a constituent with all the other arguments
({\cgf S\bs\gacc}\/ or {\cgf S\fs\gacc}), the parser
allows \gacc.
This is unavoidable unless the parser is made aware of the local and non-local
context. In other words, this method solves the spurious ambiguity problem
between higher types, but not among higher and lower types.
One can try to remedy this problem by making the availability of 
types dependent on some measures of prominence, e.g., allowing subjects
only in higher types to account for subject-complement asymmetries.
But, as pointed out by Eisner \shortcite[p.85]{eisner96}, this is not spurious
ambiguity in the technical sense, just multiple derivations due to alternative
lexical category assignments.
Eliminating ambiguity in such 
cases remains to be solved.

\section{Revealing the PAS}\label{sec:reducer}

The output of the parser is a combinatory form. The combinators in this form
may arise from the CCG schema, i.e., the compositor \combb, and 
the substitutor \combs\ \cite{steedman87}. They may also be projected
from the PAS of a lexical item, such as the duplicator 
\combw\ (with the reduction rule $\combw fa \red faa$) for reflexives,
and $\combb^{n+1}\combc$ for
predicate composition with the causative suffix.
For instance,
the combinatory form for (\ref{ex:combf}a) is the expression
(\ref{ex:combf}b).

\eenumsentence{\label{ex:combf}
\item[a.]
\begin{tabular}[t]{lll}
\exf Adam & \exf \c{c}ocu\u{g}-a & \exf kitab-\i\\
man.NOM & child-DAT & book-ACC \\
:\so{m} & :\so{c} & :\so{b} \\
\end{tabular}
\item[]\hspace*{2em}\begin{tabular}[t]{l}
 \exf oku-t-tu\\
 read-CAUS-PAST\\
 :$\combb^3\mbox{\sc caus}\combc\so{r}$\\
\end{tabular}

\centerline{'The man had the child read the book.'}

\item[b.]
$\combt\jux\so{m}\jux(\combb\jux(\combt\jux\so{b})\jux(\combt\jux\so{c})\jux
(\combb^3\jux\mbox{\sc cause}\jux\combc\jux\so{r}))\equiv$
}

\noindent\hspace*{-3em}
\leaf{\combt}\leaf{\so{m}}\branch{2}{\jux}
\leaf{\combb}\leaf{\combt}\leaf{\so{b}}\branch{2}{\jux}\branch{2}{\jux}
\leaf{\combt}\leaf{\so{c}}\branch{2}{\jux}\branch{2}{\jux}
\leaf{$\combb^3$}\leaf{\sc cause}\branch{2}{\jux}
\leaf{\combc}\branch{2}{\jux}
\leaf{\so{r}}\branch{2}{\jux}\branch{2}{\jux}\branch{2}{\jux}
\tree

Although \combb\ works in a binary manner in CCG to achieve abstraction,
it requires 3 arguments for full evaluation (its {\em order} is 3).
Revealing the PAS amounts to stripping off all combinators from the
combinatory form by evaluating the reducible expressions (redexes).
$\combb fg$ is not a redex but $\combb fga$ is. In other words,
the derivations by the parser must saturate the combinators 
in order to reveal
the PAS, which should contain no combinators. PAS is the 
{\em semantic  normal form}\/ of a derivation.

The sequence of evaluation is the normal order, which corresponds to
reducing the leftmost-outermost redex first \cite{peyton-jones87}.
In tree-theoretic terms, this is depth-first reduction of
the combinator tree in which the rearrangement is controlled by
the reduction rule of the leftmost combinator, e.g., 
$\combt\so{m}X \red X\so{m}$ where $X$ is
the parenthesized subexpression in (\ref{ex:combf}b). Reduction by
\combt\ yields:

\hspace*{-3em}
\leaf{\combb}\leaf{\combt}\leaf{\so{b}}\branch{2}{\jux}\branch{2}{\jux}
\leaf{\combt}\leaf{\so{c}}\branch{2}{\jux}\branch{2}{\jux}
\leaf{$\combb^3$}\leaf{\sc cause}\branch{2}{\jux}
\leaf{\combc}\branch{2}{\jux}
\leaf{\so{r}}\branch{2}{\jux}\branch{2}{\jux}
\leaf{\so{m}}\branch{2}{\jux}
\tree

Further reductions eventually reveal the PAS:
\begin{eqnarray}
\combb\jux(\combt\jux\so{b})\jux(\combt\jux\so{c})\jux
(\combb^3\jux\mbox{\sc cause}\jux\combc\jux\so{r})\jux\so{m} \red \\
\combt\jux\so{b}\jux(\combt\jux\so{c}\jux
(\combb^3\jux\mbox{\sc cause}\jux\combc\jux\so{r}))\jux\so{m} \red \\
\combt\jux\so{c}\jux
(\combb^3\jux\mbox{\sc cause}\jux\combc\jux\so{r})\jux\so{b}\jux\so{m} \red \\
\combb^3\jux\mbox{\sc cause}\jux\combc\jux\so{r}\jux\so{c}\jux
\so{b}\jux\so{m} \red \\
\label{rul:c1}
\mbox{\sc cause}\jux(\combc\jux\so{r}\jux\so{c}\jux 
\so{b})\jux\so{m} \red \\ 
\label{rul:c2}
\mbox{\sc cause}\jux(\so{r}\jux\so{b}\jux
\so{c})\jux\so{m}
\end{eqnarray} 

By the second Church-Rosser theorem, normal order evaluation will
terminate if the combinatory form has a normal form. But Combinatory
Logic has the same power as $\lambda-$calculus, and suffers
from the same undecidability results. For instance, 
$\combw\combw\combw$ has no normal form because the reductions will never
terminate. Some terminating reductions, such as 
$\combc\combi\combi b \red b\combi$, has no normal form either. 
It is an open question as to whether such forms can be projected from a natural
language lexicon. In an expression $X\jux Y$ where 
$X$ is not a redex, the evaluator 
recursively evaluates to reduce as much as possible 
because $X$ may contain other redexes, 
as in (\ref{rul:c1}) above. Recursion is terminated either by obtaining the
normal form, as in (\ref{rul:c2}) above, or by equivalence check.
For instance, $(\combc\jux(\combi\jux a)\jux b)\jux Y$ recurses
on the left subexpression to obtain $(\combc\jux a\jux b)$ then gives up on
this subexpression since
the evaluator returns the same expression without further evaluation.

\section{Conclusion}\label{sec:conclusion} 

If an ordered representation of the PAS is assumed as many theories
do nowadays, its derivation from the surface string 
requires that the category assignment for case cues be
rich enough in word order and 
grammatical function information to correctly place the arguments 
in the PAS. This work shows that these categories and their types 
can be uniquely characterized in the lexicon and tightly controlled in
parsing. Spurious ambiguity problem is kept under control by normal form 
parsing on the syntactic side with the use of labelled categories in the
grammar. 
Thus, the PAS of a derivation
can be determined uniquely even in the presence of type shifting.
The same strategy can account for
deriving the PAS in unbounded constructions and non-constituent
coordination \cite{bozsahin97}. 

Parser's output (the combinatory form) is reduced to a PAS
by normal order evaluation.
Model-theoretic interpretation can proceed
in parallel with derivations, or as a post-evaluation stage which takes
the PAS as input. Quantification and scrambling in free word order
languages interact in many ways,
and future work will concentrate
on this aspect of semantics.

\bibliographystyle{acl}

\begin{thebibliography}{}

\bibitem[\protect\citename{Alsina}1996]{alsina96}
Alex Alsina.
\newblock 1996.
\newblock {\em The Role of Argument Structure in Grammar}.
\newblock CSLI, Stanford, CA.

\bibitem[\protect\citename{Bozsahin and Gocmen}1995]{bozsahingocmen95}
Cem Bozsahin and Elvan Gocmen.
\newblock 1995.
\newblock A categorial framework for composition in multiple linguistic
  domains.
\newblock In {\em Proceedings of the Fourth International Conference on
  Cognitive Science of {NLP}}, Dublin.

\bibitem[\protect\citename{Bozsahin}1997]{bozsahin97}
Cem Bozsahin.
\newblock 1997.
\newblock Grammatical functions and word order in {C}ombinatory {G}rammar.
\newblock ms.

\bibitem[\protect\citename{Bresnan and Kanerva}1989]{bresnankanerva89}
Joan Bresnan and Jonni~M. Kanerva.
\newblock 1989.
\newblock Locative inversion in {C}hichewa: A case study of factorization \ in
  grammar.
\newblock {\em Linguistic Inquiry}, 20:1--50.

\bibitem[\protect\citename{Calder \bgroup et al.\egroup
  }1988]{calderkleinzeevat88}
Jonathan Calder, Ewan Klein, and Henk Zeevat.
\newblock 1988.
\newblock Unification categorial grammar.
\newblock In {\em Proceedings of the 12th International Conference on
  Computational\ Linguistics}, Budapest.

\bibitem[\protect\citename{Curry and Feys}1958]{curryfeys58}
Haskell~B. Curry and Robert Feys.
\newblock 1958.
\newblock {\em Combinatory Logic I}.
\newblock North-Holland, Amsterdam.

\bibitem[\protect\citename{Dowty}1988]{dowty88}
David Dowty.
\newblock 1988.
\newblock Type raising, functional composition, and non-constituent
  conjunction.
\newblock In Richard~T. Oehrle, Emmon Bach, and Deirdre Wheeler, editors, {\em
  Categorial Grammars and Natural Language Structures}. D. Reidel, Dordrecht.

\bibitem[\protect\citename{Eisner}1996]{eisner96}
Jason Eisner.
\newblock 1996.
\newblock Efficient normal-form parsing for combinatory categorial grammar.
\newblock In {\em Proceedings of the 34th Annual Meeting of the {ACL}}, pages
  79--86.

\bibitem[\protect\citename{Grimshaw}1990]{grimshaw90}
Jane Grimshaw.
\newblock 1990.
\newblock {\em Argument Structure}.
\newblock MIT Press, Cambridge, MA.

\bibitem[\protect\citename{Hepple and Morrill}1989]{hepplemorrill89}
Mark Hepple and Glyn Morrill.
\newblock 1989.
\newblock Parsing and derivational equivalence.
\newblock In {\em Proceedings of the 4th EACL}, Manchester.

\bibitem[\protect\citename{Hoffman}1995]{hoffman95}
Beryl Hoffman.
\newblock 1995.
\newblock {\em The Computational Analysis of the Syntax and Interpretation of
  ``Free'' Word Order in {T}urkish}.
\newblock {Ph.D.} thesis, University of Pennsylvania.

\bibitem[\protect\citename{Karttunen}1989]{karttunen89}
Lauri Karttunen.
\newblock 1989.
\newblock Radical lexicalism.
\newblock In Mark Baltin and Anthony Kroch, editors, {\em Alternative
  Conceptions of Phrase Structure}. Chicago University Press.

\bibitem[\protect\citename{{Peyton Jones}}1987]{peyton-jones87}
Simon~L. {Peyton Jones}.
\newblock 1987.
\newblock {\em The Implementation of Functional Programing Languages}.
\newblock Prentice-Hall, New York.

\bibitem[\protect\citename{Shaumyan}1987]{shaumyan87}
Sebastian Shaumyan.
\newblock 1987.
\newblock {\em A Semiotic Theory of Language}.
\newblock Indiana University Press.

\bibitem[\protect\citename{Steedman}1987]{steedman87}
Mark Steedman.
\newblock 1987.
\newblock Combinatory grammars and parasitic gaps.
\newblock {\em Natural Language and Linguistic Theory}, 5:403--439.

\bibitem[\protect\citename{Steedman}1988]{steedman88}
Mark Steedman.
\newblock 1988.
\newblock Combinators and grammars.
\newblock In Richard~T. Oehrle, Emmon Bach, and Deirdre Wheeler, editors, {\em
  Categorial Grammars and Natural Language Structures}. D. Reidel, Dordrecht.

\bibitem[\protect\citename{Steedman}1990]{steedman90}
Mark Steedman.
\newblock 1990.
\newblock Gapping as constituent coordination.
\newblock {\em Linguistics and Philosophy}, 13:207--263.

\bibitem[\protect\citename{Steedman}1996]{steedman96}
Mark Steedman.
\newblock 1996.
\newblock {\em Surface Structure and Interpretation}.
\newblock {MIT} Press, Cambridge, MA.

\bibitem[\protect\citename{Vijay-Shanker and Weir}1993]{vijay-shankerweir93}
K.~Vijay-Shanker and David~J. Weir.
\newblock 1993.
\newblock Parsing some constrained grammar formalisms.
\newblock {\em Computational Linguistics}, 19:591--636.

\bibitem[\protect\citename{Wechsler}1995]{wechsler95}
Stephen Wechsler.
\newblock 1995.
\newblock {\em The Semantic Basis of Argument Structure}.
\newblock CSLI, Stanford, CA.

\bibitem[\protect\citename{Wittenburg}1987]{wittenburg87}
Kent Wittenburg.
\newblock 1987.
\newblock Predictive combinators.
\newblock In {\em Proceedings of the 25th Annual Meeting of the {ACL}}, pages
  73--79.

\end{thebibliography}

\end{document}